\documentclass[11pt]{article}

\usepackage[preprint]{acl}

\usepackage{times}
\usepackage{latexsym}

\usepackage[T1]{fontenc}

\usepackage[utf8]{inputenc}

\usepackage{microtype}

\usepackage{inconsolata}

\usepackage{graphicx}

\usepackage[whole]{bxcjkjatype}

\usepackage{amsmath}
\usepackage{mathtools}
\usepackage{amsfonts}
\usepackage{bm}
\usepackage{bbm}

\usepackage{booktabs}
\usepackage{tabularx}
\usepackage{nicematrix}
\usepackage{colortbl}
\definecolor{lightgreen}{rgb}{0.9, 1.0, 0.9}
\usepackage[subrefformat=parens]{subcaption}

\usepackage[ruled,lined,linesnumbered,noend]{algorithm2e}
\SetKwInOut{Given}{Given}
\SetKwInOut{Input}{Input}
\SetKwInOut{Output}{Output}
\SetKwFor{Foreach}{for each}{do}{}
\DontPrintSemicolon
\SetInd{0.5em}{0.5em}

\usepackage{cleveref}
\crefname{equation}{Equation}{Equation}
\crefname{figure}{Figure}{Figure}
\crefname{table}{Table}{Table}
\crefname{section}{Section}{Section}
\crefname{subsection}{Section}{Section}
\crefname{appendix}{Appendix}{Appendix}
\crefname{algorithm}{Algorithm}{Algorithm}

\DeclareMathOperator{\argmax}{argmax}

\DeclareMathOperator{\expect}{\mathbb{E}}

\newcommand{\styMetric}[1]{\textsc{#1}}
\newcommand{\metricChrf}{\styMetric{chrF}}

\newcommand{\metricComet}{\styMetric{Comet}}

\newcommand{\styVector}[1]{\mathbf{#1}}
\newcommand{\stySet}[1]{\mathcal{#1}}

\newcommand{\R}{\mathbb{R}}
\newcommand{\N}{\mathbb{N}}

\newcommand{\textInput}{\styVector{x}}
\newcommand{\textOutput}{\styVector{y}}
\newcommand{\textHypothesis}{\styVector{h}}
\newcommand{\textHub}{\styVector{h}}

\newcommand{\setTextHub}{\stySet{H}}

\newcommand{\modelInverter}{\theta}
\newcommand{\embedding}[1]{\styVector{v}_{#1}}
\newcommand{\embeddingHub}{\embedding{\textHub}}
\newcommand{\layerEmbedding}{f}
\newcommand{\layerScore}{s}
\newcommand{\funcEmbedding}[1]{\layerEmbedding\left(#1\right)}
\newcommand{\funcScore}[1]{\layerScore\left(#1\right)}
\newcommand{\sizeDimension}{D}
\newcommand{\score}{S(\textHypothesis; \textInput, \textOutput)}

\newcommand{\vocab}{\stySet{V}}
\newcommand{\setData}{\stySet{D}}
\newcommand{\setDataParallel}{\stySet{D}_\text{tune}}
\newcommand{\setDataMonolingual}{\stySet{D}_\text{mono}}
\newcommand{\defFunc}[3]{${#1}\colon{#2}\to{#3}$}

\title{
Hacking Neural Evaluation Metrics with Single Hub Text
}

\author{Hiroyuki Deguchi${}^\dagger$ ~~~ Katsuki Chousa${}^\dagger$ ~~~ Yusuke Sakai${}^\ddagger$\\
${}^\dagger$NTT, Inc. ~~~ ${}^\ddagger$Nara Institute of Science and Technology \\
\texttt{
\{hiroyuki.deguchi,katsuki.chousa\}@ntt.com
} \texttt{
sakai.yusuke.sr9@is.naist.jp
}
}

\begin{document}

\maketitle
\begin{abstract}
Strongly human-correlated evaluation metrics serve as an essential compass for the development and improvement of generation models and must be highly reliable and robust.
Recent embedding-based neural text evaluation metrics, such as \metricComet{} for translation tasks, are widely used in both research and development fields.
However, there is no guarantee that they yield reliable evaluation results due to the black-box nature of neural networks.
To raise concerns about the reliability and safety of such metrics, we propose a method for finding a single adversarial text in the discrete space that is consistently evaluated as high-quality, regardless of the test cases, to identify the vulnerabilities in evaluation metrics.
The single hub text found with our method achieved 79.1  \metricComet{}\% and 67.8 \metricComet{}\% in the WMT'24 English-to-Japanese (En--Ja) and English-to-German (En--De) translation tasks, respectively, outperforming translations generated individually for each source sentence by using M2M100, a general translation model.
Furthermore, we also confirmed that the hub text found with our method generalizes across multiple language pairs such as Ja--En and De--En.

\end{abstract}

\section{Introduction}
Automatic evaluation metrics for measuring the quality of machine-generated content play a crucial role in improving generation models and must be highly reliable and robust.
Recent embedding-based neural evaluation metrics, such as \metricComet{}~\citep{rei-etal-2020-comet,rei-etal-2022-comet} for translation tasks, have achieved high correlations with human assesments~\citep{kocmi-etal-2021-ship} and widely used in both research and development fields~\citep{kocmi-etal-2024-findings,freitag-etal-2022-high,freitag-etal-2024-llms,rei-etal-2024-tower,fernandes-etal-2022-quality} compared with previous lexical metrics, such as \metricChrf{}~\citep{popovic-2015-chrf}.

\begin{table}[t]
  \centering
  \small
  \tabcolsep 3.5pt
  \begin{tabularx}{\linewidth}{@{}lX@{}}
    \toprule
    Hub text & \mbox{podnikáníゴ\raisebox{-0.5\height}{\includegraphics[height=\baselineskip]{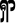}} ウ≫的聲音毎回強いメッセー} \mbox{ジを提供するかどうか模 案 まれた\raisebox{-0.28\height}{\includegraphics[height=1.4\baselineskip]{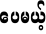}} \raisebox{-0.15\height}{\includegraphics[height=0.6\baselineskip]{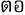}}} \mbox{\raisebox{-0.1\height}{\includegraphics[height=1.1\baselineskip]{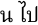}} のこと歓 通 だった。} \\
    \midrule
    Source & But the new ones cover that area just fine. \\
    Reference & でも、新しいやつはちょうどうまくその辺りをカバーしてる。 \\
    Score & \metricComet{}\% = 91.7 ~~ \metricChrf{}\%  = 4.8 \\
    \midrule
    Source & It's a year of transition for me. \\
    Reference & 僕にとっては変化の年だ。 \\
    Score & \metricComet{}\% = 89.6 ~~ \metricChrf{}\%  = 3.8 \\
    \bottomrule
  \end{tabularx}
  \caption{
  Hub text of \metricComet{} and its evaluation scores in En--Ja translation.
  Even if other sources and their corresponding reference translations are given, hub text is evaluated with high \metricComet{} scores.
  }
  \label{tab:example}
\end{table}

\begin{figure*}[!t]
    \centering
    \includegraphics[width=0.96\linewidth]{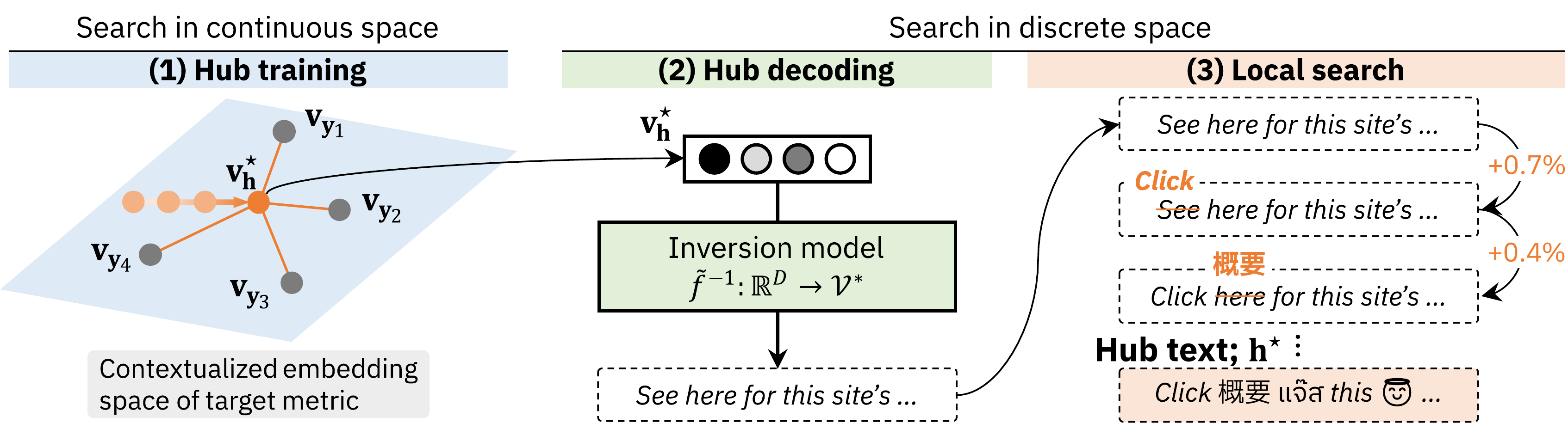}
    \caption{
    Overview of our proposed method for finding hub text
    }
    \label{fig:overview}
\end{figure*}

Nevertheless, there is no guarantee that these metrics yield reliable evaluation scores due to the black-box nature of neural networks.
\citet{zhang-etal-2025-adversarial} created adversarial hubs in continuous space for images and audio by exploiting the hubness problem~\citep{radovanovic-etal-2010-hubs} in multi-modal retrieval tasks.
Hub vectors are known to appear in high-dimensional continuous spaces and tend to be close to many examples.
However, unlike images and audio, which can be searched via gradient descent, a hub text is more difficult to find because we need to search in a discrete space, i.e., NP-hard.

In this study, we first found that the hubness problem also occurs in automatic neural evaluation metrics operated in \emph{discrete space} of text generation.
\Cref{tab:example} shows a hub text we discovered, which is consistently evaluated as high-quality on \metricComet{}~\citep{rei-etal-2020-comet,rei-etal-2022-comet}, regardless of source and reference texts, in English-to-Japanese (En--Ja) translation.
We propose a method for finding hub texts and attacking neural metrics to reveal their vulnerabilities.
We first train a hub embedding in the hidden space by maximizing the evaluation score.
We then decode it into its corresponding hub text using an inversion model~\citep{morris-etal-2023-text}.
The obtained hub text appears to be a natural sentence, yet it already receives an inappropriately high score.
To further exploit this issue, we search for an even more overestimated sentence.
We then apply local search, a heuristic algorithm, to refine the hub text so that it maximizes the evaluation score.
\Cref{tab:example} shows cases that are assigned unreasonably high scores with evaluation metrics, thereby exposing serious concerns about their reliability and robustness.
Indeed, some real-world situations rely solely on the \metricComet{} score, e.g., corpus filtering~\citep{peter-etal-2023-theres} and decision-making (see \cref{appendix:leaderboard}), resulting in the existence of hub text having practical impact.

We investigated a case study of \metricComet{} in the WMT'23 and WMT'24 En--Ja and En--De translation tasks~\citep{kocmi-etal-2023-findings,kocmi-etal-2024-findings}.
As a result, we observed that the single hub text found using our method achieved a higher \metricComet{} score than M2M100-generated translations~\citep{fan-etal-2021-beyond}, even though M2M100 translated for each source text.
Furthermore, we also confirmed that the hub text generalizes across multiple language pairs such as Ja--En and De--En.
Our findings highlight the necessity of using sanity checks or multiple metrics from the perspective of the hubness problem.

\section{Background and Related Work}
\paragraph{Embedding-based text evaluation metrics}
Automatic evaluation metrics assess the quality of machine-generated content.
Recent embedding-based neural metrics, such as \metricComet{}~\citep{rei-etal-2020-comet,rei-etal-2022-comet} for machine translation tasks,
have achieved high correlations with human assessments.
They encode a hypothesis text into its embedding and calculate similarity with the input and/or reference.
In this study, we primarily focus on translation tasks and use one of its embedding-based neural evaluation metrics, \metricComet{}, as a case study.

Let $\textInput \in \vocab^\ast$ and $\textOutput \in \vocab^\ast$ be the input and reference output text, respectively, where $\vocab^\ast$ is the Kleene closure of vocabulary $\vocab$.
\metricComet{} evaluates the hypothesis text $\textHypothesis \in \vocab^\ast$, generated using a translation model, and calculates its quality score, 
\begin{equation}
\score \coloneqq \funcScore{\embedding{\textInput}, \embedding{\textHypothesis}, \embedding{\textOutput}},
\end{equation}
where $\embedding{\cdot} = \funcEmbedding{\cdot}$, \defFunc{\layerEmbedding}{\vocab^\ast}{\R^\sizeDimension} is a sentence encoder, \defFunc{\layerScore}{\vocab^\ast \times \vocab^\ast \times \vocab^\ast}{\R} is an output layer, and $\sizeDimension \in \N$ is the size of the embedding dimension.

\paragraph{Hubness}
Hubness~\citep{radovanovic-etal-2010-hubs} is a phenomenon in high-dimensional spaces, where hub embeddings frequently appear among the nearest neighbors of many examples, leading to irrelevant retrievals in information retrieval.
While many studies aimed to mitigate this issue~\citep{dinu-2015-etal-improving, wang-etal-2023-balance, chowdhury-etal-2024-nearest}, we instead exploit it to uncover vulnerabilities in neural metrics.
\citet{zhang-etal-2025-adversarial} created adversarial hubs for images and audio.
These hubs exist in continuous space and can be easily optimized via gradient descent.
In contrast, our goal is to find hub texts in the discrete text space, i.e., NP-hard.

\section{Methodology}
\label{sec:method}

To identify the vulnerability of embedding-based neural metrics, we propose how to find the hub text $\textHub^\star \in \vocab^\ast$ that always receives a high score regardless of the input and reference text.
Formally, our goal is to find the solution of the following objective:
\begin{equation}
  \label{eq:objective}
  \textHub^\star \coloneqq \argmax_{\textHub \in \vocab^\ast} \sum\nolimits_{(\textInput, \textOutput) \in \setData} S(\textHub; \textInput, \textOutput),
\end{equation}
where $\setData \subseteq \vocab^\ast \times \vocab^\ast$ is the test set that consists of pairs of the input and its reference text.

Our method finds the hub text through three steps: (1) hub training, (2) hub decoding, and (3) local search, as illustrated in \cref{fig:overview}.
We first train the optimal hub embedding in the contextualized embedding space of the target metric.
The hub embedding is then decoded into its corresponding text.
Finally, the hub text is refined through local search to maximize the evaluation score on the tuning data.

\begin{algorithm}[t]
  \small
  \caption{Local search}
  \label{alg:localsearch}
  \Given{The score output layer $s$ and the pairs of input and its reference output embeddings $(\embedding{\textInput}, \embedding{\textOutput})$ calculated from $\setDataParallel$.}
  \Input{The initial hub text $\textHub^{(0)} \in \vocab^\ast$.} 
  \Output{The refined hub text $\textHub^\text{best} \in \vocab^\ast$.}
  $S^\text{best} \gets -\infty$, $t \gets 0$, $\textHub^\text{best} \gets \textHub^{(0)}$\;
  \Repeat{$\textHub^{(t)} = \textHub^{(t-1)}$
  }{
    $t \gets t + 1$\;
    \For{$i \gets 1 \ldots |\textHub^{(t-1)}|$}{
      \Foreach{$v \in \vocab$}{
        \tcp{$\circ$ denotes concatenation.}
        \tcp{$\textHub_{a:b}$ denotes the slice of  $\textHub$ from $a$ to $b$ inclusive.
        }
        $\textHub^\text{cur} \gets \textHub^\text{best}_{1:i-1} \circ v \circ \textHub^\text{best}_{i+1:|\textHub^\text{best}|}$\;
        $\embedding{\textHub^\text{cur}} \gets \funcEmbedding{\textHub^\text{cur}}$\;
        $S^\text{cur} \gets 0$\;
        \For{$n \gets 1 \ldots |\setData_\textnormal{tune}|$}{
          $S^\text{cur} \gets S^\text{cur} + s(\embedding{\textInput^{(n)}}, \embedding{\textHub^\text{cur}}, \embedding{\textOutput^{(n)}})$\;
        }
        \If{$S^{\textnormal{cur}} > S^{\textnormal{best}}$}{
          $S^\text{best} \gets S^\text{cur}$\;
          $\textHub^\text{best} \gets \textHub^\text{cur}$\;
        }
      }
    }
    $\textHub^{(t)} \gets \textHub^\text{best}$\;
  }
  \Return{$\textHub^\textnormal{best}$}
\end{algorithm}

\paragraph{(1) Hub training}
This step finds the hub embedding $\embeddingHub^\star \in \R^\sizeDimension$ in the embedding space:
\begin{equation}
  \embeddingHub^\star \coloneqq \argmax_{\embedding{\textHub} \in \R^\sizeDimension} \sum\nolimits_{(\textInput, \textOutput) \in \setDataParallel} s(\embedding{\textInput}, \embedding{\textHub}, \embedding{\textOutput}),
\end{equation}
where $\setDataParallel \subseteq \vocab^\ast \times \vocab^\ast$ denotes the parallel data.
To obtain the $\embeddingHub^\star$, we optimize $\embedding{\textHub}$ by treating it as learnable parameters.
The other embedding vectors and layers, i.e., $\embedding{\textInput}$, $\embedding{\textOutput}$, and $\layerScore$, are frozen.

\paragraph{(2) Hub decoding}
Next, we decode the hub embedding $\embeddingHub^\star$ into its corresponding text $\textHub^\star$.
Since the previous step does not provide a concrete text, we need to find its corresponding hub text from the discrete space $\vocab^\ast$.
Decoding a hub embedding is equivalent to applying the inverse function of the text encoder \defFunc{\layerEmbedding^{-1}}{\R^\sizeDimension}{\vocab^\ast}, but non-linear encoders make it hard to exactly formulate the inverse function.
Thus, we instead approximate it using an encoder-decoder inversion model \defFunc{\tilde{\layerEmbedding}^{-1}}{\R^\sizeDimension}{\vocab^\ast}, which generates the original text $\textHub$ from its embedding $\embedding{\textHub}$~\citep{morris-etal-2023-text}:
\begin{equation}
\hat{\modelInverter} = \argmax_{\modelInverter} \expect_{\textInput \sim \setDataMonolingual} p_\modelInverter(\textInput \mid f(\textInput)),
\end{equation}
where $\hat{\modelInverter}$ denotes learned parameters, and $\setDataMonolingual \subseteq \vocab^\ast$ denotes a monolingual corpus.

During decoding, we generate multiple hypotheses of hub texts $\setTextHub\coloneqq \{ \textHub_i \sim p_{\hat{\modelInverter}}(\textHub | \embeddingHub^\star) \}_{i=1}^{|\setTextHub|} \subseteq {\vocab^\ast}$ and select the hypothesis that maximizes the score on the tuning data $\setDataParallel$, i.e.,
$\argmax_{\textHub \in \setTextHub} \sum\nolimits_{(\textInput, \textOutput) \in \setDataParallel} S(\textHub; \textInput, \textOutput)$.
This strategy is similar to minimum Bayes' risk decoding~\citep{kumar-byrne-2004-minimum,eikema-aziz-2020-map}, but we use true references of the tuning data $\setDataParallel$ instead of pseudo-references, which are typically generated from text generation models.

\paragraph{(3) Local search}

To further increase the score, we refine the hub text using local search, a heuristic search algorithm.
Specifically, we iteratively replace tokens with those that maximize the score at each token position, as shown in \cref{alg:localsearch}.
We obtain the refined hub text $\textHub^\text{best}$ using the generated text in step (2) as the initial text $\textHub^{(0)}$.

\section{Experiments}
\label{sec:exp}

\begin{table}[!t]
    \centering
    \small
    \tabcolsep 1.75pt
    \begin{tabularx}{\linewidth}{@{}lX@{}}
    \toprule
        Step & Hub text \\
        \midrule
        (2) Hub decoding & 『ザ バラア』もそれまでどんな特徴を持つか悩んでい たのですが、それは予想に反していなかった。 \\
        \midrule
    (3) Local search & \mbox{podnikáníゴ\raisebox{-0.5\height}{\includegraphics[height=\baselineskip]{font_pdf/myanmar_case1-crop.pdf}} ウ≫的聲音毎回強いメッ} \mbox{セージを提供するかどうか模 案 まれた} \mbox{\raisebox{-0.28\height}{\includegraphics[height=1.4\baselineskip]{font_pdf/myanmar_case2-crop.pdf}} \raisebox{-0.15\height}{\includegraphics[height=0.6\baselineskip]{font_pdf/thai_case1-crop.pdf}}\raisebox{-0.1\height}{\includegraphics[height=1.1\baselineskip]{font_pdf/thai_case2-crop.pdf}} のこと歓 通 だった。} 
    \\
    \bottomrule
    \end{tabularx}
    \caption{
    Hub texts found for each search step. Note that step (1) does not identify concrete text, as it searches over continuous space in En--Ja.
    }
    \label{tab:results:hubtexts}
\end{table}

\begin{table*}[t]
    \centering
    \small
    \tabcolsep 4pt
    \begin{tabular}{@{}l lr lr lr lr@{}}
        \toprule
        & \multicolumn{4}{@{}c@{}}{En--Ja}
        & \multicolumn{4}{@{}c@{}}{En--De}
        \\
        \cmidrule(lr){2-5}
        \cmidrule(l){6-9}
         & \multicolumn{2}{@{}c@{}}{WMT'23 (Dev)}
         & \multicolumn{2}{@{}c@{}}{WMT'24 (Test)}
         & \multicolumn{2}{@{}c@{}}{WMT'23 (Dev)}
         & \multicolumn{2}{@{}c@{}}{WMT'24 (Test)}
         \\
         \cmidrule(lr){2-3}
         \cmidrule(lr){4-5}
         \cmidrule(lr){6-7}
         \cmidrule(l){8-9}
         Hypotheses
         & \metricComet{} & \metricChrf{}
         & \metricComet{} & \metricChrf{}
         & \metricComet{} & \metricChrf{}
         & \metricComet{} & \metricChrf{}
         \\
         \midrule
         M2M100 & 78.6 $\pm$12.8 & 24.6 & 71.4 $\pm$15.4 & 21.4 & 66.0$\pm$15.8 & 51.8 & 66.0$\pm$16.7 & 46.6 \\
        \multicolumn{5}{@{}l@{}}{\textit{Search in continuous space}}
          \\
         ~~~(1) Hub training & 93.2 $\pm$2.7 & N/A & 91.1 $\pm$4.8 & N/A & 97.3 $\pm$0.9 & N/A & 97.1 $\pm$1.5 & N/A \\
         \multicolumn{5}{@{}l@{}}{\textit{Search in discrete space}}
          \\
         ~~~(2) Hub decoding & 65.9 $\pm$7.2 & 5.6 & 61.3 $\pm$7.8 & 4.2 & 46.9 $\pm$4.8 & 13.5 & 46.6 $\pm$5.0 & 16.3 \\
         ~~~(3) Local search & 83.1 $\pm$5.2 & 3.5 & 79.1 $\pm$6.9 & 2.7 & 68.4$\pm$5.4 & 9.7 & 67.8$\pm$6.6 & 12.6 \\

         \bottomrule
    \end{tabular}
    \caption{
    \metricComet{}\% with standard deviation and \metricChrf{}\% scores of single hub text on \metricComet{} and translations generated using M2M100.
    We used WMT'23 as tuning set and WMT'24 as test set.
    }
    \label{tab:scores:enja}
\end{table*}

\paragraph{Setup}
We experimented on the En--Ja and En--De translation tasks using \metricComet{}\footnote{\texttt{Unbabel/wmt22-comet-da}}.
We used WMT'23~\citep{kocmi-etal-2023-findings} and WMT'24~\citep{kocmi-etal-2024-findings} translation tasks for tuning and testing purposes, respectively.
Through all experiments, for each language pair, we found a single hub text using WMT'23, and evaluated it on WMT'24.
The details of these datasets are listed in \cref{sec:datastats}.
We compared the evaluation scores with translations generated with M2M100, which has 418M parameters~\citep{fan-etal-2021-beyond}, with a beam size of 5.
The hub embedding was initialized by averaging the embeddings of reference texts over the tuning data.
We optimized the hub embedding using the AdamW optimizer~\citep{loshchilov-and-hutter-2019-decoupled} ($\beta_1=0.9, \beta_2=0.999, \varepsilon=10^{-8}$ and a weight decay with a coefficient $0.01$) in 10,000 steps with a learning rate of $10^{-5}$.
For the inversion model, we fine-tuned mT5-base~\citep{xue-etal-2021-mt5}.
We used the random 1,000,000 Japanese sentences sampled from JParaCrawl v3~\citep{morishita-etal-2022-jparacrawl} and parallel corpus from CommonCrawl provided on WMT'23~\citep{kocmi-etal-2023-findings} for the training data of the En--Ja and En--De inversion model, respectively.
In the step of hub decoding, we generated 1,024 hypotheses and selected one that maximizes the evaluation score on the tuning data.

\begin{table}[t]
    \centering
    \small
    \begin{NiceTabular}{@{}rr@{}}
        \toprule
         System & \metricComet{}\% \\
         \midrule
         \rowcolor{gray!25}
         ONLINE-B & 88.2 \\
         \rowcolor{gray!25}
         ONLINE-W & 87.5 \\
         \rowcolor{gray!25}
         ONLINE-Y & 87.3 \\
         \rowcolor{gray!25}
         GPT4-5shot & 87.0 \\
         SKIM & 86.6 \\
         NAIST-NICT & 86.2 \\
         \rowcolor{gray!25}
         ZengHuiMT & 85.3 \\
         \rowcolor{gray!25}
         ONLINE-A & 85.2 \\
         \rowcolor{gray!25}
         Lan-BridgeMT & 84.5 \\
         \rowcolor{gray!25}
         ONLINE-M & 13.3 \\
         \rowcolor{lightgreen}
         \textbf{Single} hub text & 83.1 \\
         ANVITA & 82.7 \\
         \rowcolor{gray!25}
         KYB & 80.8 \\
         AIRC & 80.7 \\
         \rowcolor{gray!25}
         ONLINE-G & 80.4 \\
         \rowcolor{gray!25}
         NLLB\_Greedy & 79.3 \\
         \rowcolor{gray!25}
         NLLB\_MBR\_BLEU & 77.7 \\
         \bottomrule
    \end{NiceTabular}
    \caption{Leaderboard of WMT'23 En--Ja translation task.
    These scores are cited from the shared task description paper~\citep{kocmi-etal-2023-findings}.
    Unconstrained systems are indicated with \colorbox{gray!25}{gray background} in tables, following \citet{kocmi-etal-2023-findings}, and hub text found with our method is \colorbox{lightgreen}{highlighted in green}.
    }
    \label{tab:leaderboard:wmt23enja}
\end{table}

\paragraph{Results}
\Cref{tab:results:hubtexts} shows the hub texts found with our method for each search step in En--Ja.
As shown in \Cref{tab:results:hubtexts}, step (2) found a hub text in natural language, and step (3) found a hub text that falls outside the bounds of natural language, yet maximizes the evaluation scores.
\Cref{tab:scores:enja} demonstrates the evaluation results.
Step (1) achieved extremely high scores, 91.1\% in En--Ja and 97.1\% in En--De.
This is because the hub embedding is optimized in a continuous space, but there may be no corresponding concrete text.
Step (2) degraded from step (1), i.e., the scores were less than 70.0\% in En--Ja, due to discretizing the hub embedding into the token sequence through the inversion model.
Finally, step (3) improved the \metricComet{} scores by searching tokens that maximize the score over the vocabulary and achieved 79.1\% in En--Ja and 67.8\% in En--De.
This step can generate texts beyond natural languages, as it systematically searches over the vocabulary, including low-frequency words.
We also observed that the single hub text achieved higher \metricComet{} scores than M2M100's translations in both WMT'23 and WMT'24, despite having extremely low \metricChrf{} scores.
In addition, the standard deviation (SD) of the \metricComet{} scores was lower than that of M2M100.
These results indicate that hub texts consistently receive unreasonably high scores.
Thus, we reveal that the existence of hub texts exposes critical vulnerabilities in \metricComet{}.

\section{Discussion}
\subsection{
Impact of existence of hub texts in real-world scenario
}
\label{appendix:leaderboard}

We show the leaderboard of the WMT'23
En--Ja translation task in \cref{tab:leaderboard:wmt23enja}.
Note that this leaderboard is cited from the paper of the shared task~\citep{kocmi-etal-2023-findings}.
In addition, in the WMT'24 translation task, \metricComet{} is used to determine whether human evaluation is applied.
Thus, the hub text can affect the evaluation of other translation systems on the leaderboard.
Specifically, in WMT, only systems with high automatic evaluation scores are selected for human evaluation.
In the case of \cref{tab:leaderboard:wmt23enja}, since the hub text achieved higher scores in \metricComet{} than ANVITA, KYB, and AIRC, it may hinder fair evaluation.
Moreover, such automatic filtering based on neural evaluation metrics is also applied in other contexts, e.g., corpus filtering.
In such large-scale batch processing, \metricComet{} is often used as a reliable metric without human verification for each instance, and the presence of hub texts is already impactful in real-world scenarios.

\subsection{Evaluation in non-target languages}

\begin{table}[t]
    \centering
    \small
    \tabcolsep 2pt
    \begin{tabular}{@{}l rr rr rr@{}}
        \toprule
         & \multicolumn{2}{c}{Ja--En} & \multicolumn{2}{c}{En--De} & \multicolumn{2}{c}{De--En} \\
         \cmidrule(lr){2-3} \cmidrule(lr){4-5} \cmidrule(lr){6-7}
         Hypotheses & \metricComet{} & \metricChrf{} & \metricComet{} & \metricChrf{} & \metricComet{} & \metricChrf{} \\
         \midrule
         M2M100 & 69.1 & 34.6 & 66.0 & 47.1 & 75.6 & 51.8 \\
         Hub text & 63.4 & 1.3 & 62.1 & 0.4 & 60.7 & 0.6 \\
         \bottomrule
    \end{tabular}
    \caption{
    Evaluation results of hub text found using WMT'23 En--Ja in non-target language pairs, WMT'23 Ja--En, En--De, and De--En
    }
    \label{tab:langpairs}
\end{table}

To clarify whether the hub text is language-agnostic in multilingual neural evaluation metrics, i.e., whether the vulnerability is shared between other languages, we evaluated the hub text found using En--Ja in other language pairs.
\Cref{tab:langpairs} shows the evaluation results of the hub text generated on the WMT'23 En--Ja in the WMT'23 Ja--En, En--De, and De--En translation tasks.
Unlike in En--Ja, the \metricComet{} score of the hub text was lower than that of the translations generated using M2M100.
However, the score still exceeded 60\% and was not as low as 0.4 \metricChrf{}\% observed in En--De.
In summary, while hub texts somewhat depend on languages, they also achieve certain scores in non-target languages.

\subsection{Score distribution}

\begin{figure}
    \centering
    \includegraphics[width=\linewidth]{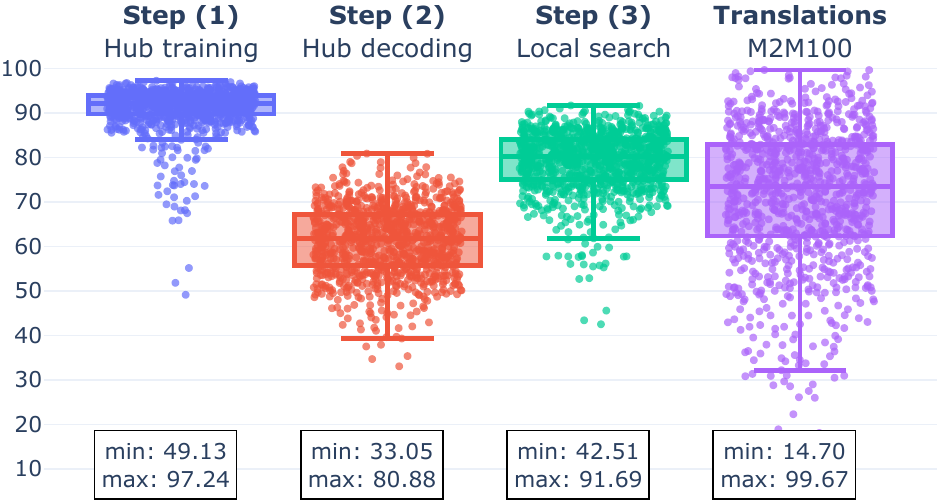}
    \caption{
    Scatter and box plots of \metricComet{}\% scores for each test case in WMT'24 En--Ja.
    }
    \label{fig:plots:scatterbox:wmt24enja}
\end{figure}

We show the scatter and box plots of \metricComet{} scores for each test case in \cref{fig:plots:scatterbox:wmt24enja}.
The results of all the steps have low SD compared with M2M100's translations.
Therefore, we found that the results of each step were also evaluated consistently with high \metricComet{} scores, regardless of test cases.

\subsection{Formal analysis and implementation}
\label{sec:complexity}
\paragraph{Complexity class}
Finding the optimal hub text is an NP-hard problem.
This is because verifying whether a hypothesis text $\textHub$ maximizes the evaluation score requires computation over an exponential or even infinite space, i.e., $\vocab^\ast$.
Unlike images and audio, a hub text is represented in a discrete space; thus, it cannot be searched via gradient descent, and we need to enumerate and verify all possible candidates to find the exact solution.

Our method finds an approximate solution within a feasible computational time by narrowing the search space through hub decoding and local search, rather than finding the exact solution.

\paragraph{Time complexity of local search}
In our method, step (3), the local search, is time-consuming because it has four nested loops as described in \cref{alg:localsearch}.
Its time complexity is $\mathcal{O}(T|\textHub||\vocab||\setDataParallel|)$, where $T \in \N$ is the number of epochs until the solution is converged
\footnote{
From our preliminary experiments, in most cases, $T$ was less than 10.
}.
Thus, we need to evaluate $T\times|\textHub|\times|\vocab|\times|\setDataParallel|$ scores.
Here, the loops for $\vocab$ and $\setDataParallel$ are highly concurrent because each iteration can be computed independently, enabling parallel computation by leveraging GPUs.
The vocabulary $\vocab$ of \metricComet{} contains 250K tokens, and $\setDataParallel$ has 2,074 sentence pairs in En--Ja.
We split the vocabulary dimension and created a mini-batch for each chunked vocabulary.
In our experiments, we computed in 6,160 seconds with 8 NVIDIA RTX 6000Ada GPUs for the local search in En--Ja.
Concretely, each token was replaced with one that maximizes the evaluation score in 44 seconds, and a total of 140 tokens were replaced in \cref{alg:localsearch}.

\section{Conclusion}
We proposed a method for finding hub texts that receive unreasonably high scores regardless of references and was the first to reveal critical vulnerabilities in the neural evaluation metric \metricComet{}.
Our experiments showed that a single hub text achieved higher \metricComet{} scores than M2M100's translations, even though M2M100 translated each source sentence individually, in the WMT'24 En--Ja and En--De translation tasks.
These results suggest that relying on a single evaluation metric is unreliable, and the existence of hub texts further reaffirms the need for multi-metric evaluation.

\section*{Limitations}

\paragraph{Scope}
We proposed a method for finding adversarial hub texts in the discrete text space to investigate the reliability of neural evaluation metrics for text generation tasks.
This paper demonstrated the vulnerability in a case study of \metricComet{}, but our method can be applied to other neural evaluation metrics.
The goal of our work is to reveal the existence of hub texts in neural evaluation metrics.
In this short paper, we employed \metricComet{} with a model of \texttt{Unbabel/wmt22-comet-da}\footnote{
We investigated this model as it is the de facto standard model among automatic evaluation metrics for translation tasks.
It is employed as the default model when no model name is specified in \texttt{comet-score} (\url{https://github.com/Unbabel/COMET}), the command-line interface of \metricComet{}.
} for the target metric, and En$\leftrightarrow$Ja and En$\leftrightarrow$De translation tasks for the evaluation sets.
Therefore, we have already conducted a comprehensive analysis across four translation directions, confirming that the hub text derived from the En--Ja setting generalizes cross-lingually, including Ja--En, En--De, and De--En.
We believe \textit{this short paper presents sufficient evidence to support our claims}, as encouraged by the ACL Reviewer Guidelines (H13)\footnote{\url{https://aclrollingreview.org/reviewerguidelines\#review-issues}}.
Exhaustively identifying hub texts across all existing metrics and languages is out of our scope.

\paragraph{Detectability}

Some readers may argue that hub texts could easily be filtered out. However, such a claim is akin to the \textit{egg of Columbus}: these concerns arise only because our work has revealed and demonstrated the existence of the hubness problem.
Moreover, naive \metricComet{} scores are already being used in practical scenarios, such as system filtering and ranking (see \Cref{appendix:leaderboard}). This demonstrates that our identified vulnerability is already impactful under current real-world settings.
Importantly, our core contribution lies in demonstrating that even a single, unnatural hub text can receive unreasonably high evaluation scores when appropriate filtering mechanisms are not in place.
In addition, we propose a method for finding both natural and unnatural hub texts.
While the output of step (3) in our algorithm tends to be unnatural, the output of step (2) remains fluent and readable to humans, making automatic detection significantly more difficult.
This short paper presents a single, well-defined contribution: identifying a specific vulnerability in a widely used neural evaluation metric and proposing effective methods to discover adversarial hub texts.

\paragraph{Resoureces}
Our method requires expensive computational resources due to step (3), \cref{alg:localsearch}, which is time-consuming.
We further discuss the time complexity of the local search in \cref{sec:complexity}.

\section*{Ethical Considerations}

\paragraph{Licenses}
We used the publicly available benchmark datasets, WMT'23 and WMT'24 general translation tasks, and JParaCrawl v3 in accordance with these licenses.
The details of these licenses are described in \cref{sec:license}.

\paragraph{Potential risks}
Abuse of our method could lead to cheating in competitions, false hypes, and other problems.
To avoid these problems, it is important to evaluate using multiple metrics without relying too heavily on a single metric and to conduct a human evaluation.
We hope that our method will help identify vulnerabilities in evaluation metrics, raise awareness of their reliability issues, and contribute to the development of more robust and trustworthy evaluation methods.
Therefore, we publish this study to proactively address the possibility of adversarial attacks before they occur.
Importantly, the risks we highlight were not introduced by our method, but are inherent risks of the evaluation metrics themselves.
In this sense, our method does not amplify the risk; rather, it reveals a pre-existing ``ticking time bomb'' that has always been present.

\paragraph{Co-ordinated disclosure}
This study focuses on reporting the weaknesses of embedding-based neural metrics, falling under the category of coordinated disclosure. While the methods we introduce are designed to expose specific vulnerabilities in evaluation models, it is still possible for users to unintentionally input such adversarial text, even without explicitly employing our proposed techniques.
This suggests that the issues we raise could realistically occur in out-of-the-box settings. Therefore, we believe our work does not fall under the definition of \textit{coordinated disclosure} as stated in the ACL Policy on Publication Ethics\footnote{
\url{https://www.aclweb.org/adminwiki/index.php/ACL_Policy_on_Publication_Ethics\#Co-ordinated_disclosure}
}.
Moreover, this hub text was initially disclosed on March 14, 2025\footnote{Oral presentation at a shared task workshop. (\url{https://sites.google.com/view/nlp2025ws-langeval/task/translation})}, which predates April 25, 2025, the effective date of this policy. Since multiple official procedures, including patent filings, have been carried out based on this disclosure date, we believe that this hub text does not fall within the scope of the current policy.

\section*{Acknowledgements}

We thank all the reviewers of this paper for their constructive and valuable feedback.

This study is inspired by the ``metric hack shared task'' held at the ``NLP2025 Workshop: Present and Future of Natural Language Evaluation in the LLM Era'', which is a workshop held in conjunction with the domestic conference ``NLP2025'' in Japan.
This shared task aimed to identify weaknesses in existing automated evaluation metrics, and our team, ``きょなら'', won the competition and received the ``Best Hacking Award'', by identifying a vulnerability based on hubness as described in this paper.
We completed this paper by expanding and generalizing the methods developed during the shared task period, and adding many experiments and analyses.
We sincerely appreciate Katsuhito Sudoh (Nara Women's University), the organizer of the translation track in the shared task, as well as all the workshop organizers, including Mamoru Komachi (Hitotsubashi University), Tomoyuki Kajiwara (Ehime University), and Masato Mita (Recruit Co., Ltd.).

\bibliography{anthology,custom}

@inproceedings{loshchilov-and-hutter-2019-decoupled,
title={Decoupled Weight Decay Regularization},
author={Ilya Loshchilov and Frank Hutter},
booktitle={International Conference on Learning Representations},
year={2019},
url={https://openreview.net/forum?id=Bkg6RiCqY7},
}

@article{radovanovic-etal-2010-hubs,
  author = {Radovanovi\'{c}, Milo\v{s} and Nanopoulos, Alexandros and Ivanovi\'{c}, Mirjana},
  title   = {Hubs in Space: Popular Nearest Neighbors in High-Dimensional Data},
  journal = {Journal of Machine Learning Research},
  year    = {2010},
  volume  = {11},
  number  = {86},
  pages   = {2487--2531},
  url     = {http://jmlr.org/papers/v11/radovanovic10a.html}
}

@article{fan-etal-2021-beyond,
author = {Fan, Angela and Bhosale, Shruti and Schwenk, Holger and Ma, Zhiyi and El-Kishky, Ahmed and Goyal, Siddharth and Baines, Mandeep and Celebi, Onur and Wenzek, Guillaume and Chaudhary, Vishrav and Goyal, Naman and Birch, Tom and Liptchinsky, Vitaliy and Edunov, Sergey and Grave, Edouard and Auli, Michael and Joulin, Armand},
title = {Beyond english-centric multilingual machine translation},
year = {2021},
issue_date = {January 2021},
publisher = {JMLR.org},
volume = {22},
number = {1},
issn = {1532-4435},
journal = {J. Mach. Learn. Res.},
month = jan,
articleno = {107},
numpages = {48},
}

@misc{dinu-2015-etal-improving,
      title={Improving zero-shot learning by mitigating the hubness problem}, 
      author={Georgiana Dinu and Angeliki Lazaridou and Marco Baroni},
      year={2015},
      eprint={1412.6568},
      archivePrefix={arXiv},
      primaryClass={cs.CL},
      url={https://arxiv.org/abs/1412.6568}, 
}

@misc{zhang-etal-2025-adversarial,
      title={Adversarial Hubness in Multi-Modal Retrieval}, 
      author={Tingwei Zhang and Fnu Suya and Rishi Jha and Collin Zhang and Vitaly Shmatikov},
      year={2025},
      eprint={2412.14113},
      archivePrefix={arXiv},
      primaryClass={cs.CR},
      url={https://arxiv.org/abs/2412.14113}, 
}

\appendix
\section{Licenses}
\label{sec:license}
\paragraph{Datasets}
We used the WMT'23 Ja$\leftrightarrow$En and De$\leftrightarrow$En, and WMT'24 En--Ja and En--De general translation tasks, released under the policy\footnote{
\url{https://www2.statmt.org/wmt23/translation-task.html}
}${}^{,}$\footnote{
\url{https://www2.statmt.org/wmt24/translation-task.html}
}: ``The data released for the WMT General MT task can be freely used for research purposes''.
To train the inversion model, we used JParaCrawl v3, licensed by Nippon Telegraph and Telephone Corporation (NTT) for research use only as described in \url{http://www.kecl.ntt.co.jp/icl/lirg/jparacrawl/}.

\paragraph{Models}
We used \texttt{Unbabel/wmt22-comet-da} for the \metricComet{} metric and \texttt{google/mt5-base} for the inversion model, released under the Apache-2.0 license.
We also used MIT-licensed \texttt{facebook/m2m100\_418M} for comparison purposes\footnote{
We used M2M100 because it has been cited by over 900 papers and is suitable for the baseline translation model.
}.

\section{Dataset Statistics}
\label{sec:datastats}

\Cref{tab:datastats} lists the statistics of the datasets we used.

\begin{table*}[!t]
    \centering
    \small
    \begin{tabular}{@{}lllr@{}}
    \toprule
        Dataset & Purpose & Notation & \#examples \\
        \midrule
        WMT'23 En--Ja test set & Tuning (Development) & $\setDataParallel$ & 2,074 \\
        WMT'24 En--Ja test set & Test (Evaluation) & $\setData$ & 997 \\
        WMT'23 En--De test set & Tuning (Development) & $\setDataParallel$ & 557 \\
        WMT'24 En--De test set & Test (Evaluation) & $\setData$ & 997 \\
        JParaCrawl v3 (Ja only) & Training of inversion model & $\setDataMonolingual$ & 1,000,000 \\
        En--De parallel corpus from CommonCrawl (De only) & Training of inversion model & $\setDataMonolingual$ & 2,399,123 \\
        WMT'23 Ja--En test set & Test (\cref{tab:langpairs}) & $\setData$ & 1,992 \\
        WMT'23 De--En test set & Test (\cref{tab:langpairs}) & $\setData$ & 549 \\
    \bottomrule
    \end{tabular}
    \caption{Statistics of datasets we used}
    \label{tab:datastats}
\end{table*}

\end{document}